\documentclass[runningheads]{llncs}

\usepackage[mobile]{eccv}

\usepackage{graphicx}
\usepackage{subcaption}
\usepackage{booktabs}
\usepackage{tabularx}
\usepackage{multirow}
\usepackage{makecell}
\usepackage{array}
\usepackage{siunitx}
\usepackage{wrapfig}
\usepackage{subcaption}
\usepackage{algorithm}
\usepackage{algorithmic}
\usepackage{xcolor}   
\usepackage{pifont}   
\usepackage{colortbl} 

\usepackage[accsupp]{axessibility}

\definecolor{codebg}{RGB}{248,249,251}
\definecolor{codeframe}{RGB}{220,225,232}
\definecolor{codekw}{RGB}{30,74,153}
\definecolor{codecm}{RGB}{110,115,125}
\definecolor{codestr}{RGB}{12,108,98}

\usepackage[most]{tcolorbox}

\lstdefinestyle{pyclean}{
  language=Python,
  basicstyle=\ttfamily\scriptsize,
  keywordstyle=\color{codekw}\bfseries,
  commentstyle=\color{codecm}\itshape,
  stringstyle=\color{codestr},
  numbers=none,
  showstringspaces=false,
  breaklines=true,
  breakatwhitespace=true,
  tabsize=2,
  xleftmargin=0.6em,
  xrightmargin=0.2em,
  aboveskip=0pt,
  belowskip=0pt,
  lineskip=0.0pt
}

\newtcblisting{wrapcode}{
  listing only,
  listing options={style=pyclean},
  colback=codebg,
  colframe=codeframe,
  boxrule=0.4pt,
  arc=2.2mm,
  left=1.2mm,
  right=1.2mm,
  top=1.0mm,
  bottom=1.0mm,
  enhanced,
}

\definecolor{iccvblue}{rgb}{0.21,0.49,0.74}
\usepackage{hyperref}
\hypersetup{
  colorlinks,
  citecolor=iccvblue,
  linkcolor=blue,
  urlcolor=iccvblue}

\definecolor{custompurple}{HTML}{7030A0} 
\definecolor{customred}{HTML}{C00000}    
\definecolor{customgreen}{HTML}{548235}  

\newcommand{\cmark}{\textcolor{green!50!black}{\ding{51}}}
\newcommand{\xmark}{\textcolor{red!50!black}{\ding{55}}}

\usepackage{eccvabbrv}

\usepackage[accsupp]{axessibility}  

\usepackage{hyperref}
\usepackage{orcidlink}

\begin{document}

\title{Pathwise Test-Time Correction for Autoregressive Long Video Generation} 


\author{
\textbf{Xunzhi Xiang\textsuperscript{1,2,*}},
\textbf{Zixuan Duan\textsuperscript{1,*}},
\textbf{Guiyu Zhang\textsuperscript{3}},
\textbf{Haiyu Zhang\textsuperscript{2}},
\textbf{Zhe Gao\textsuperscript{1}},
\textbf{Junta Wu\textsuperscript{2}},
\textbf{Shaofeng Zhang\textsuperscript{4}},
\textbf{Tengfei Wang\textsuperscript{2}\textsuperscript{\textdagger}},
\textbf{Qi Fan\textsuperscript{1}\textsuperscript{\textdagger}},
\textbf{Chunchao Guo\textsuperscript{2}}
\\
\small
\textsuperscript{1}Nanjing University \quad
\textsuperscript{2}Tencent Hunyuan \quad
\textsuperscript{3}Chinese University of Hong Kong, Shenzhen \quad
\textsuperscript{4}University of Science and Technology of China
\\
\small
\textsuperscript{*}Equal Contribution \quad
\textsuperscript{\textdagger}Corresponding Authors
\\
\small
\href{https://ttc-1231.github.io/}{Project Page}
}

\authorrunning{Pathwise Test-Time Correction for
Autoregressive Long Video Generation}

\institute{}

\maketitle

\begin{center}
  \includegraphics[width=\textwidth]{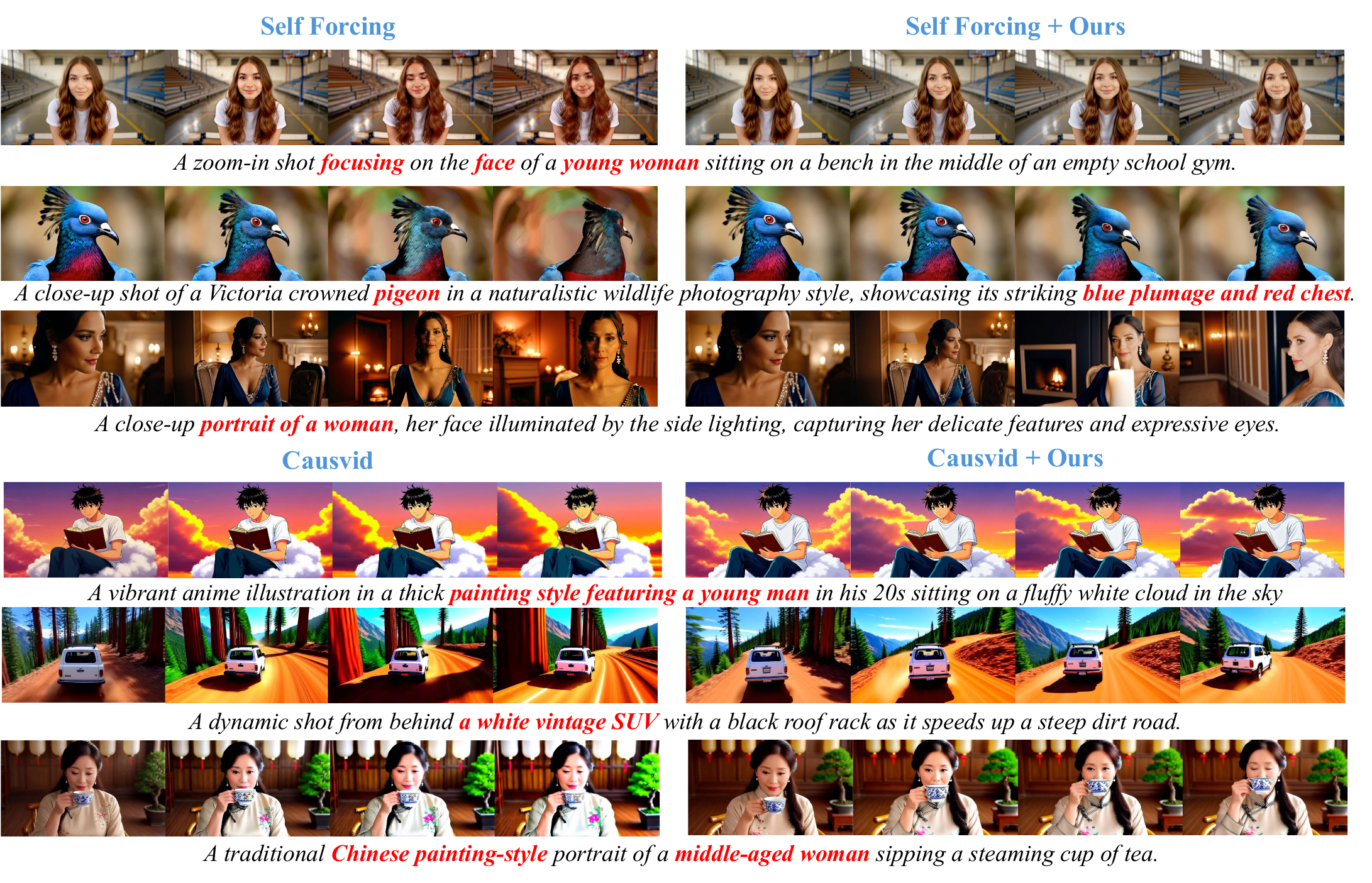}
  \captionof{figure}{\textbf{30-second video generation examples.}
  Our method reduces error accumulation in CausVid and Self-Forcing, enabling longer and more stable videos with improved visual consistency.
  All samples are generated with the \textbf{same random seed} for fair comparison.}
  \label{fig:teaser}
\end{center}

\begin{abstract}
  Distilled autoregressive diffusion models facilitate real-time short video synthesis but suffer from severe error accumulation during long-sequence generation. While existing Test-Time Optimization (TTO) methods prove effective for images or short clips, we identify that they fail to mitigate drift in extended sequences due to unstable reward landscapes and the hypersensitivity of distilled parameters. To overcome these limitations, we introduce Test-Time Correction (TTC), a training-free alternative. Specifically, TTC utilizes the initial frame as a stable reference anchor to calibrate intermediate stochastic states along the sampling trajectory. Extensive experiments demonstrate that our method seamlessly integrates with various distilled models, extending generation lengths with a slight overhead while matching the quality of resource-intensive training-based methods on 30-second benchmarks. 
  \keywords{Video Generation\and Training-Free \and Test-Time Correction}
\end{abstract}

\section{Introduction}
\label{sec:intro}
Video generation \cite{rewardforcing, infinityrope, Cogvideo, moga1} has advanced rapidly with the development of diffusion-based generative models~\cite{kong2024hunyuanvideo, wan2025,Cogvideo, Latte, DiT, LDM}, which now enable the high-quality synthesis of complex motion \cite{champ, Animateanyone} and visual appearance \cite{animatediff, Proteus}. However, scaling these diffusion priors to extended video sequences remains a formidable challenge. Beyond the escalating computational costs associated with longer contexts, maintaining temporal coherence over extended horizons is difficult without incurring excessive latency, thereby limiting their deployment in real-time applications.


\begin{wrapfigure}{r}{0.48\columnwidth}
  \vspace{-8pt}  
  \includegraphics[width=0.46\columnwidth]{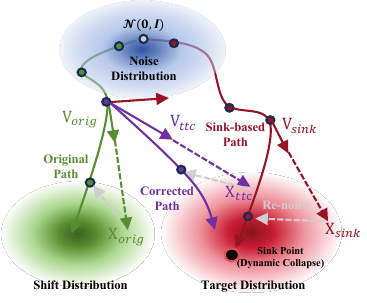}
  \caption{\textbf{Comparison of sampling strategies.} 
  The \textbf{\textcolor{customgreen}{Original Path}} suffers from error accumulation, 
  while the \textbf{\textcolor{customred}{Sink-based Path}} collapses into a \textit{Sink Point}. 
  Our \textbf{\textcolor{custompurple}{TTC strategy}} avoids this via reference-conditioned 
  denoising and explicit \textit{Re-noising}.}
  \label{fig:distribution}
  \vspace{-20pt} 
\end{wrapfigure}

To overcome these limitations, recent studies~\cite{yin2025slow, huang2025self} have shifted from bidirectional modeling to step-distilled autoregressive generation, enabling true real-time video synthesis. However, these methods remain constrained by cascading error accumulation: since each frame is conditioned on prior outputs, initial inaccuracies compound over time, resulting in temporal drift and long-horizon degradation.
While recent extensions~\cite{yi2025deep, lol} like Rolling Forcing~\cite{liu2025rolling}, LongLive~\cite{yang2025longlive}, Self-Forcing\text{++}~\cite{Self-Forcing++} , and WorldPlay~\cite{sun2025worldplay} have achieved minute-level consistency through sink mechanisms and windowed DMD retraining, they necessitate substantial computational overhead for model fine-tuning. 
Consequently, a pivotal question arises: \textit{Can we improve the stability of autoregressive video generation purely at inference time, bypassing the need for retraining the base model?}

TTO~\cite{wang2025test,sun2020test} has emerged as a compelling alternative for enhancing video quality without the need for retraining. However, while effective for short-video synthesis~\cite{yu2025autorefiner, eyring2025noise}, our toy experiments reveal that scaling TTO to long-horizon autoregressive generation faces a dual bottleneck consisting of the inherent difficulty in defining reward functions for long-range consistency and the extreme optimization sensitivity of distilled models. We observe that in these distilled models, even infinitesimal test-time gradients often trigger reward collapse and fail to mitigate cumulative error. Therefore, we propose TTC, which is a training-free framework that shifts the paradigm from parameter-space optimization to sampling-space stochastic intervention. TTC is grounded in the insight that few-step distilled samplers are inherently stochastic as they perturb intermediate states with injected noise. This property implies that intermediate predictions are not fixed outcomes but rather malleable latent states that can be rectified by subsequent diffusion steps to align with the global initial context while preserving the underlying sampling distribution.

Specifically, as shown in Figure~\ref{fig:distribution}, TTC applies a small number of correction steps along the stochastic sampling path, \textbf{only after the global structure has stabilized}. This delay prevents the generation from falling into sink-collapse~\cite{lol}, a phenomenon where newly generated frames repeatedly regress toward the sink frames instead of evolving naturally.

At chosen steps in the sampling path, TTC performs reference-conditioned denoising by utilizing the initial frame context to anchor a corrected clean prediction. Then, this corrected state is re-noised back to the variance level corresponding to the current timestep, which ensures that the intervention remains compatible with the expected noise distribution. 
By integrating correction into the stochastic sampling path of the autoregressive diffusion process rather than directly replacing the denoised prediction, this mechanism suppresses long-term error accumulation and temporal drift without retraining, while preserving high-fidelity temporal coherence over extended durations.

In this work, we show that long-horizon stability in autoregressive video generation can be achieved through test-time intervention alone. Our method suppresses error accumulation with a slight computational overhead, without requiring any retraining. As a result, it extends the stable generation length of distilled autoregressive models from a few seconds to over 30 seconds, while achieving visual quality comparable to state-of-the-art training-based methods. Consistent improvements across multiple model architectures demonstrate that TTC is a robust and general solution for stabilizing distilled autoregressive diffusion models.

\section{Related Work}

\noindent \textbf{Bidirectional Models for Video Generation.}
Diffusion models have been widely adopted for video generation, with recent works typically formulating video synthesis as a sequence-level joint denoising problem. Under this bidirectional diffusion paradigm, all frames are denoised simultaneously via spatiotemporal attention, enabling the model to leverage global temporal context and produce temporally coherent, high-fidelity videos~\cite{blattmann2023align,yin2023nuwa,jia2025moga,ma2025tempomaster,zhang2025fast, Voyager}. Large-scale systems such as Hunyuan~\cite{kong2024hunyuanvideo} and Wan~\cite{wan2025} further demonstrate the effectiveness of this joint denoising formulation at scale. However, because the entire sequence must be processed as a whole during inference, this paradigm inherently precludes streaming or incremental generation, limiting its applicability in real-time and interactive scenarios.

\begin{figure*}[t!]
  \hfill
  \includegraphics[width=\linewidth]{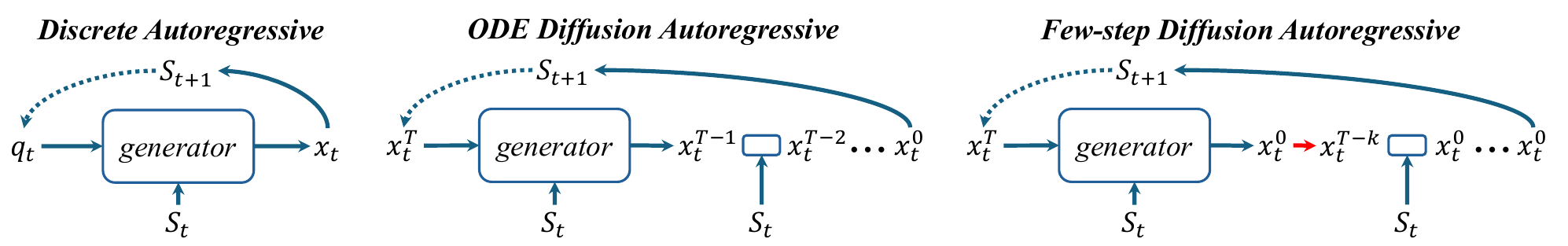}
  \caption{\textbf{Variants of autoregressive video generation.} Discrete AR uses single-step deterministic prediction, multi-step diffusion follows a deterministic ODE trajectory, while few-step distilled diffusion performs stochastic sampling with intermediate noise injection.}
  \label{fig:samples}
\end{figure*}

\noindent \textbf{Autoregressive Models for Video Generation.} Autoregressive video diffusion models generate videos sequentially in a strict causal manner, conditioning each new frame or segment on the historical context of previously generated content~\cite{chen2024diffusion,yin2025slow,huang2025self,liu2025rolling,yang2025longlive,chen2025skyreels,ji2025memflow,teng2025magi,NOVA, resampleforcing}. While this formulation naturally supports streaming generation with low initial latency, it is inherently susceptible to error accumulation, where minor deviations propagate and amplify across steps, leading to severe temporal drift and degraded coherence in long videos. To mitigate this issue, recent works introduce planning methods~\cite{zhang2025framepack,xiang2025macro} or explicit memory mechanisms~\cite{sun2025worldplay,chen2025teleworld, contex, mixtureof, memoryforcing, hyworld2025}, strategies that typically necessitate complex architectural modifications and extensive re-training.

\noindent \textbf{Test-time Image/Video Generation.}
Test-time generation methods~\cite{wang2025test,sun2020test,liang2025comprehensive} aim to enhance the performance of pre-trained models directly during the inference phase. Test-time scaling improves quality by iteratively searching over multiple candidates, as seen in Video-T1~\cite{liu2025video} and EvoSearch~\cite{he2025scaling}, yet this comes at the price of prohibitive computational cost. Similarly, Test-time optimization refines generation via auxiliary parameter updates that necessitate instance-specific training, such as HyperNoise~\cite{eyring2025noise}, AutoRefiner~\cite{yu2025autorefiner}, and SLOWFAST-VGEN~\cite{hong2025slowfastvgen}. In contrast, our approach distinguishes itself from both paradigms by being fully training-free, avoiding both the overhead of candidate search and the complexity of parameter optimization.

\section{Test-time Optimization for Distilled Models}

\subsection{Background: Few-step Distilled Sampling}
In this section, we formulate autoregressive video generation as next-chunk prediction under a context-conditional generative model.
Given a video sequence $\{x_1,\dots,x_N\}$, the joint distribution factorizes as
\begin{equation}
p(x_{1:N})=\prod_{t=1}^{N} p_\theta(x_t\mid S_t),\quad
S_t=\{x_1,\dots,x_{t-1}\},
\end{equation}
where $S_t$ denotes the context at step $t$, consisting of all previously generated frames or chunks. As illustrated in Figure~\ref{fig:samples}, existing autoregressive video generation methods typically fall into three categories.
Discrete autoregressive models generate each chunk through a single deterministic prediction conditioned on past outputs, while multi-step autoregressive diffusion models approximate the same conditional distribution via a deterministic ODE-based sampling trajectory.
In contrast, few-step distilled diffusion models replace deterministic ODE solvers with a \emph{stochastic sampling process} that explicitly injects noise at intermediate steps.

Under this formulation, conditional distribution $p_\theta(x_t\mid S_t)$ is no longer realized as a single deterministic mapping, but through a \emph{stochastic diffusion sampling trajectory} defined over a sparse set of diffusion steps
$\{T_0=0,T_1,\dots,T_K=T_{\max}\}$.
Specifically, distilled video generation begins from Gaussian noise $x_t^{T_{\max}}\sim\mathcal{N}(0,I)$
and evolves progressively along this trajectory via a sequence of denoise--renoise transitions.
Specifically, at each denoising step $T_j$, the generation process starts from a noisy latent state and applies the denoising network to produce an estimate of the underlying clean latent representation,
\begin{equation}
x_{t,0}^{\,T_j}
=
G_\theta\!\left(
\Psi(x_{t,0}^{\,T_{j+1}},\epsilon_t^{\,T_j},T_j);\,
S_t,T_j
\right),
\end{equation}
where $G_\theta(\cdot)$ denotes the parameterized denoising network, and $S_t$ represents the autoregressive context at step $t$.

After each denoising step, distilled diffusion models proceed by re-injecting noise according to the predefined schedule, mapping the clean estimate back onto the diffusion trajectory. Concretely, the estimated clean latent is re-noised to obtain the latent state at the next diffusion step,
\begin{equation}
x_t^{\,T_{j-1}}
=
\Psi\!\left(
x_{t,0}^{\,T_j},\epsilon_t^{\,T_{j-1}},T_{j-1}
\right),\quad
\epsilon_t^{\,T_{j-1}}\sim\mathcal{N}(0,I),
\end{equation}
thereby yielding a stochastic transition that advances the generation process to the next noise level. 

The forward diffusion process $\Psi(\cdot)$ is defined as
\begin{equation}
\Psi(x,\epsilon_T,T)=\alpha_T\,x+\sigma_T\,\epsilon_T,
\end{equation}
where $\alpha_T$ and $\sigma_T$ are predefined diffusion coefficients corresponding to step $T$. Repeating this denoise–re-noise procedure across diffusion steps forms the complete stochastic sampling trajectory of distilled diffusion models.

\begin{figure}[t]
\centering

\begin{minipage}{0.48\columnwidth}
    \centering
    \includegraphics[width=\linewidth]{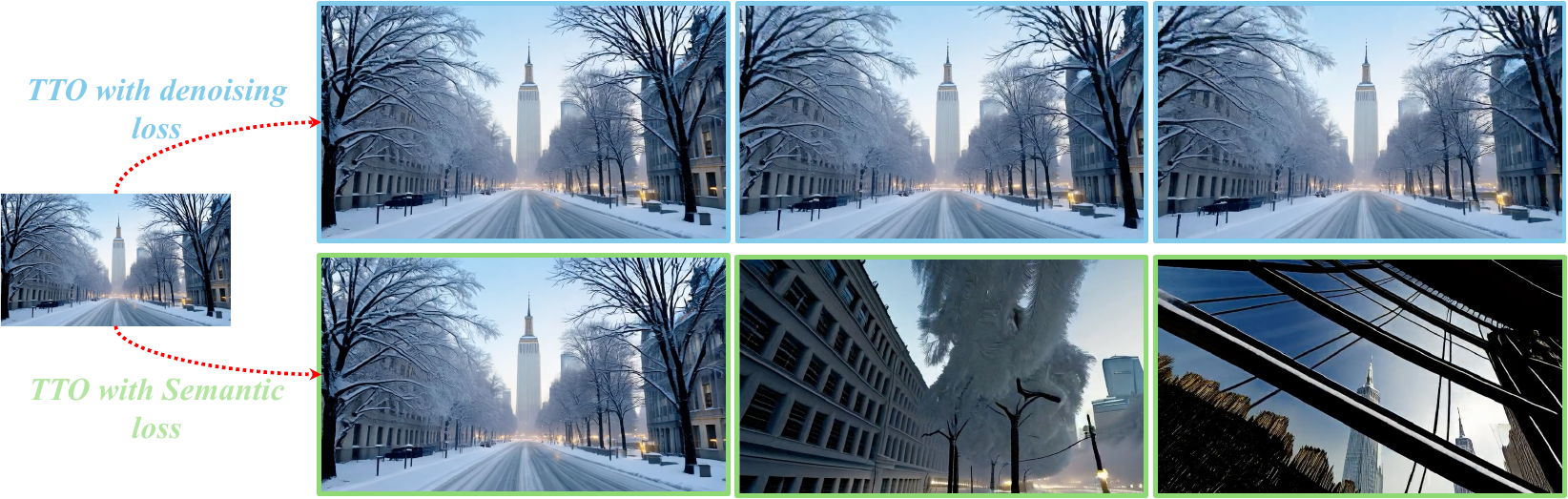}
    \captionof{figure}{Comparison of two toy test-time optimization variants based on LoRA fine-tuning, illustrating distinct failure modes.}
    \label{fig:toy}
\end{minipage}
\hfill
\begin{minipage}{0.48\columnwidth}
    \centering
    \includegraphics[width=\linewidth]{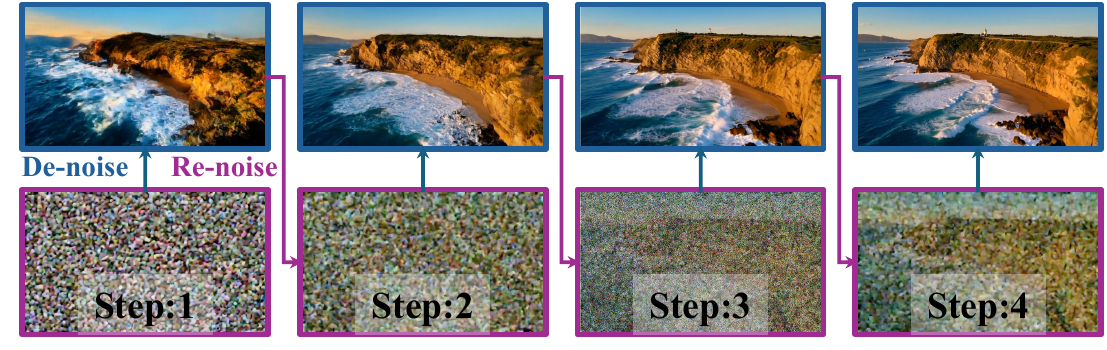}
    \captionof{figure}{\textbf{Intermediate predictions along the stochastic sampling path.} High-noise steps determine global structure, while low-noise steps refine appearance details under a fixed layout.}
    \label{fig:noise_regime}
\end{minipage}

\end{figure}

\subsection{Toy Experiment: Applying Test-time Optimization to Long Video Generation}
Existing TTO methods improve generation quality by aligning the model distribution with a predefined reward function.
Given a pre-trained generative model with output distribution $p^{\mathrm{base}}$, TTO typically defines a reward-weighted target distribution
\begin{equation}
p^*(x) \propto p^{\mathrm{base}}(x)\exp(r(x)),
\end{equation}
where $r(x)$ encodes preferences of samples $x$.
The optimization objective can be formulated as minimizing the KL divergence between a parameterized distribution $p^\phi$ and the target distribution $p^*$,
\begin{equation}
\min_\phi D_{\mathrm{KL}}(p^\phi \| p^*)
= \min_\phi D_{\mathrm{KL}}(p^\phi \| p^{\mathrm{base}})
- \mathbb{E}_{x \sim p^\phi}[r(x)],
\end{equation}
which trades off reward maximization against deviation from the original model distribution.
However, for long video generation, it remains challenging to design an explicit reward that effectively suppresses error accumulation. Temporal drift arises from coupled inconsistencies in semantics, appearance, and motion, which are difficult to characterize with a single hand-crafted objective. A naive alternative is to constrain each subsequent chunk’s predictive distribution to remain close to that of the initial frames, effectively anchoring generation to early content.

To assess this idea, we conduct two toy experiments using direct LoRA fine-tuning~\cite{LoRA} at test time, following HyperNoise~\cite{eyring2025noise} and AutoRefiner~\cite{yu2025autorefiner}.
Both variants use the same backbone and identical LoRA adapters, and differ only in their optimization objectives.
The first variant fine-tunes LoRA with a standard denoising reconstruction loss on early frames across noise levels.
The second variant replaces pixel-level reconstruction with a semantic consistency objective, enforcing similarity to early frames in pretrained feature spaces~\cite{CLIP, dinov2}.

Figure~\ref{fig:toy} shows that these two objectives lead to distinct failure modes.
The reconstruction-based variant quickly collapses to a trivial solution, where later frames become near-duplicates of the initial frame, resulting in severe motion loss.
In contrast, the semantic objective fails to effectively reduce long-horizon error accumulation, and the generated videos still exhibit temporal drift similar to the baseline.
These results indicate that naive TTO, whether based on low-level reconstruction or high-level semantics, is insufficient for stable long-horizon generation.

\begin{figure*}[t]
  \hfill
  \centering
  \includegraphics[width=\linewidth]{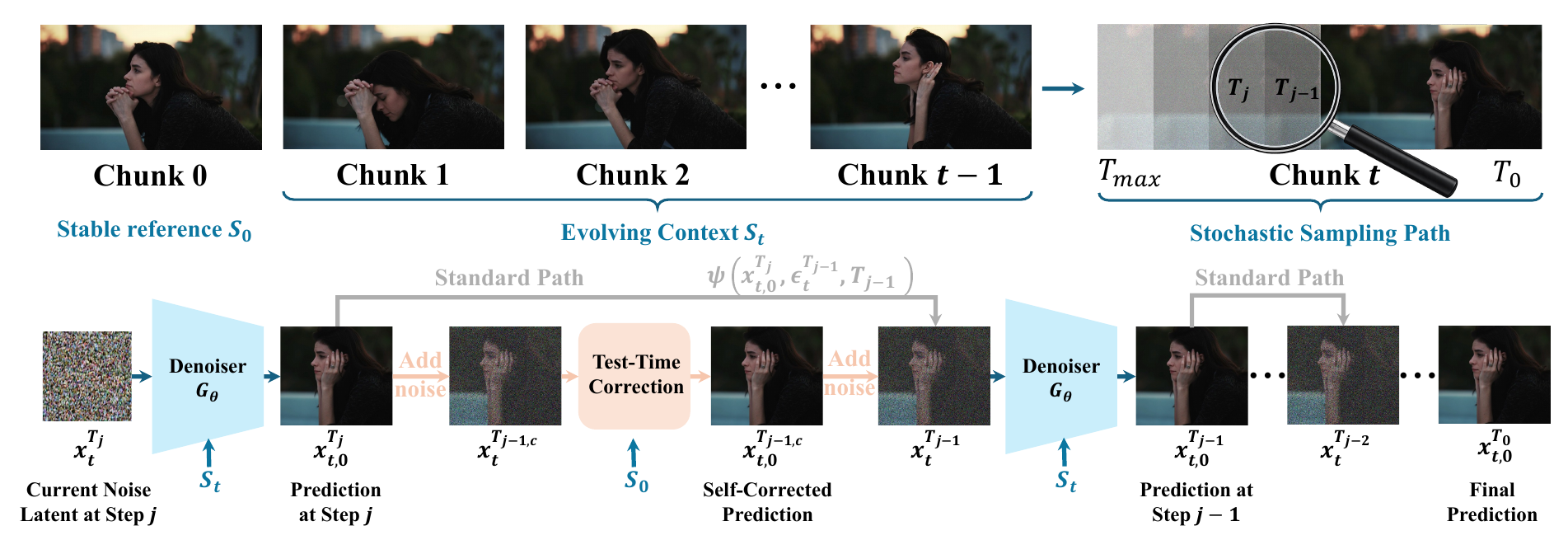}
  \caption{\textbf{Overall pipeline of our method.} A sparse set of correction steps is inserted into the stochastic sampling path until the global structure stabilizes. At selected steps, TTC performs reference-conditioned denoising using the initial frame to obtain a corrected prediction, which is then re-noised to the current timestep to remain  consistent with the expected noise distribution. This on-path, training-free correction suppresses long-term error accumulation and stabilizes long-horizon generation.}
  \label{fig:our_method}
\end{figure*}

\section{From Test-Time Optimization to Test-Time Correction}
Based on the above analysis, we identify two key limitations of TTO for long video generation.

\noindent \textbf{Reward design for error accumulation.}
Temporal drift stems from coupled errors in semantics, appearance, and motion, which are hard to capture with a single reward: low-level reconstruction suppresses motion, while high-level semantic objectives lack frame-wise correction signals.

\noindent \textbf{Optimization Challenges and Collapse.} Performing test-time optimization on distilled models presents significant training difficulties. The models tend to overfit rapidly to the auxiliary reward, causing the optimization trajectory to collapse into specific, degenerate solutions that violate the pre-trained generative prior.

Together, these limitations motivate a shift from parameter-updating test-time optimization to \emph{test-time correction}, which avoids model updates and instead performs trajectory-aware interventions during sampling.


\subsection{Correctability along the Stochastic Sampling Path}
\label{sec:self_correctability}
Distilled few-step diffusion models maintain a stochastic sampling trajectory through iterative noise re-injection, which prevents premature convergence and preserves flexibility in intermediate states. As shown in Figure~\ref{fig:noise_regime}, this trajectory exhibits a clear functional phase transition. At high noise levels, the denoising process primarily determines global structure, such as scene layout and spatial relationships. As the noise level decreases, the generation progressively shifts to an appearance refinement stage, where local textures and fine visual details are synthesized while the global structure remains largely fixed.
This phase-wise behavior naturally suggests a principled test-time correction strategy. Rather than intervening throughout the sampling process, we apply alternative conditioning only during the appearance refinement stage, after the global structure has stabilized. At this stage, the model is less sensitive to structural changes, allowing visual attributes to be adjusted without affecting layout or geometry. As a result, targeted test-time intervention can modulate appearance while preserving structural consistency.

Motivated by planning-based video generation models~\cite{zhang2025framepack, xiang2025macro}, which relax strictly unidirectional prediction via cross-frame context, we consider applying test-time intervention at a \emph{single sampling step} after structural stabilization. Specifically, at a designated step $j^\star$, we restrict the visible context state $S_t$ to include only the earliest frame, forcing the model to rely exclusively on the initial frame for subsequent appearance refinement and texture generation. This \textbf{single-point correction} process can be formalized as:

\begin{equation}
x_{t,0}^{\,T_{j-1}}
=
G_\theta\!\left(
\Psi(x_{t,0}^{\,T_{j}}, \epsilon_t^{\,T_{j-1}}, T_{j-1});\,
S_t^{(j \rightarrow j^\star)},\,
T_{j-1}
\right),
\end{equation}
where $S_t^{(j \rightarrow j^\star)}$ denotes the modified context state. Specifically, at the designated sampling step $j^\star$, the original autoregressive context $S_t$ is replaced by the earliest-frame context $S_0$, while all other sampling steps remain unchanged. Here, $j^\star$ corresponds to the stage at which the global layout and object structure have stabilized.

\subsection{Pathwise Test-time Correction}
\label{sec:pathwise_tto}

Despite its conceptual simplicity, single-point latent correction frequently leads to visible artifacts in distilled autoregressive diffusion models, such as flickering, abrupt appearance changes, and temporal inconsistency. To overcome this, we propose a path-wise self-correction strategy as shown in Figure~\ref{fig:our_method}, which leverages the model's stochastic nature to ensure smooth state transitions.

Instead of performing a hard correction at a single denoising step, our method first applies the intervention to the current prediction and then re-noises it back to the current noise level. By restarting the sampling process from this re-noised state, the correction is naturally integrated into the stochastic path. This avoids the abrupt state transitions typical of direct prediction replacement, allowing the model to smoothly assimilate the update while maintaining generation stability.

Formally, consider generating chunk $t$ under a denoising schedule $T_{\max} > \cdots > T_j > T_{j-1} > \cdots > T_0$. At step $j$, given the current noisy latent $x_t^{T_j}$ and the evolving context $S_t$, the denoiser produces a clean prediction as:
\begin{equation}
x_{t,0}^{T_j}
= G_\theta\!\left(x_t^{T_j};\, S_t,\, j\right).
\label{eq:pred_j}
\end{equation}
Instead of directly executing the next denoising update following the standard path in Figure~\ref{fig:our_method}, we first apply forward diffusion noise injection to the current prediction and explicitly map it to the next noise level $T_{j-1}$. Then, we replace the evolving context $S_t$ with a stable reference context $S_0$ for denoising. This produces a reference-aligned corrected clean prediction as:
\begin{equation}
x_{t,0}^{T_{j-1},c}
= G_\theta\!\left(
\Psi\!\left(x_{t,0}^{T_j},\, \epsilon_t^{j-1},\, T_{j-1}\right);
\, S_0,\, j-1
\right),
\label{eq:corr_j}
\end{equation}

The corrected prediction is then mapped back to the same noise level $T_{j-1}$ through noise injection. Denoising is subsequently resumed under the true evolving context $S_t$
\begin{equation}
x_{t,0}^{T_{j-1}}
= G_\theta\!\left(
\Psi\!\left(x_{t,0}^{T_{j-1},c},\, \tilde{\epsilon}_t^{j-1},\, T_{j-1}\right);
\, S_t,\, j-1
\right),
\label{eq:resume_j}
\end{equation}
This sequence of operations integrates test-time correction directly into the stochastic sampling process. All modified intermediate states are produced through valid diffusion transitions, as summarized in Algorithm~\ref{alg:pathwise_tto}. This effectively suppresses chunk-boundary flickering, mitigates long-horizon error accumulation, and preserves temporal coherence in autoregressive video generation.

\begin{algorithm}[t]
\caption{Pathwise Test-time Correction}
\label{alg:pathwise_tto}
\begin{algorithmic}[1]
\newcommand{\algG}[1]{\textcolor{RoyalBlue}{#1}}       
\newcommand{\algPsi}[1]{\textcolor{Violet}{#1}}        
\newcommand{\algRef}[1]{\textcolor{BrickRed}{#1}}      
\newcommand{\algCtx}[1]{\textcolor{ForestGreen}{#1}}   
\newcommand{\algSet}[1]{\textcolor{teal}{#1}}          

\newcommand{\algComment}[1]{\hfill \textcolor{Gray}{\footnotesize\ttfamily \# #1}} 
\newcommand{\algSection}[1]{\STATE \textit{\textcolor{darkgray}{#1}}} 
\STATE \textbf{Input:} Noise schedule $\{T_J > \cdots > T_0 = 0\}$; Generator $\algG{G_\theta}$; Evolving context $\algCtx{S_t}$; Ref context $\algRef{S_0}$; Correction indices $\algSet{\mathcal{J}^\star}$; Diffusion forward process $\algPsi{\Psi}$
\STATE \textbf{Output:} Final prediction $x_{t,0}^{0}$

\STATE Sample $x_t^{T_J} \sim \mathcal{N}(0, I)$ \algComment{Initial gaussian noise}

\FOR{$j = J$ \textbf{down to} $1$}
    \STATE $x_{t,0}^{T_j} \leftarrow \algG{G_\theta}\!\left(x_t^{T_j};\, \algCtx{S_t},\, j\right)$ \algComment{Initial prediction with \algCtx{$S_t$}}
    \STATE Sample $\epsilon_t^{T_{j-1}} \sim \mathcal{N}(0, I)$

    \IF{$j-1 \in \algSet{\mathcal{J}^\star}$}
        \algSection{--- Phase A: Reference-guided Correction ---}
        \STATE $x_t^{T_{j-1},c} \leftarrow \algPsi{\Psi}\!\left(x_{t,0}^{T_j},\, \epsilon_t^{T_{j-1}},\, T_{j-1}\right)$ 
        
        \STATE $x_{t,0}^{T_{j-1},c} \leftarrow \algG{G_\theta}\!\left(x_t^{T_{j-1},c};\, \algRef{S_0},\, j-1\right)$ \algComment{Correct trajectory using \algRef{$S_0$}}
        
        \algSection{--- Phase B: Re-noising \& Re-denoising ---}
        \STATE Sample $\tilde{\epsilon}_t^{T_{j-1}} \sim \mathcal{N}(0, I)$
        \STATE $x_t^{T_{j-1}} \leftarrow \algPsi{\Psi}\!\left(x_{t,0}^{T_{j-1},c},\, \tilde{\epsilon}_t^{T_{j-1}},\, T_{j-1}\right)$ \algComment{Inject new gaussian noise}
        \STATE $x_{t,0}^{T_{j-1}} \leftarrow \algG{G_\theta}\!\left(x_t^{T_{j-1}};\, \algCtx{S_t},\, j-1\right)$ \algComment{Finalize step with \algCtx{$S_t$}}
    \ELSE
        \STATE $x_t^{T_{j-1}} \leftarrow \algPsi{\Psi}\!\left(x_{t,0}^{T_j},\, \epsilon_t^{T_{j-1}},\, T_{j-1}\right)$
        \STATE $x_{t,0}^{T_{j-1}} \leftarrow \algG{G_\theta}\!\left(x_t^{T_{j-1}};\, \algCtx{S_t},\, j-1\right)$
    \ENDIF
\ENDFOR
\STATE \textbf{return} $x_{t,0}^{T_0}$
\end{algorithmic}
\end{algorithm}

\begin{table}[t]
\scriptsize
\centering
\caption{\textbf{Comprehensive comparison with SOTA methods on prompt-conditioned 30-second video generation (Speed \& VBench).}}
\label{tab:merged_speed_vbench}
\setlength{\tabcolsep}{1.5pt} 

\resizebox{\textwidth}{!}{%
\begin{tabular}{l c c cccccc}
\toprule
\multirow{2}{*}{\textbf{Method}} &
\multirow{2}{*}{\makecell[c]{\textbf{Training}\\\textbf{Free}}} &
\textbf{Speed} &
\multicolumn{6}{c}{\textbf{VBench Metrics $\uparrow$}} \\
\cmidrule(lr){3-3} \cmidrule(lr){4-9}
& & \makecell[c]{Total\\fps} &
\makecell[c]{Subject\\Consistency} &
\makecell[c]{Background\\Consistency} &
\makecell[c]{Dynamic\\Degree} &
\makecell[c]{Motion\\Smoothness} &
\makecell[c]{Imaging\\Quality} &
\makecell[c]{Aesthetic\\Quality} \\
\midrule
Rolling Forcing  & \xmark & 15.38 & 95.8 & 95.1 & 35.9 & 98.9 & 72.5 & 63.6 \\
LongLive         & \xmark &  -    & 95.5 & 95.4 & 44.5 & 98.8 & 71.7 & 65.0 \\
CausVid          & \xmark & 15.79 & 91.2 & 91.4 & 50.8 & 98.1 & 70.2 & 63.5 \\
\rowcolor{gray!10} \textbf{CV + Ours} & \cmark & 10.53 & 93.2 & 93.3 & 69.5 & 97.6 & 70.1 & 63.5 \\
Self-Forcing     & \xmark & 15.79 & 92.5 & 93.2 & 62.5 & 98.0 & 72.5 & 63.4 \\
\rowcolor{gray!10} \textbf{SF + Ours} & \cmark & 10.53 & 94.0 & 94.2 & 60.2 & 98.3 & 72.7 & 63.8 \\
\bottomrule
\end{tabular}%
}
\end{table}

\begin{table}[htbp]
\scriptsize
\centering
\setlength{\tabcolsep}{4pt}
\renewcommand{\arraystretch}{1.15}
\caption{\textbf{Comprehensive comparison with SOTA methods on prompt-conditioned 30-second video generation (Color-shift \& JEPA).}
We report Color-shift metrics (L1, Correlation) and JEPA consistency (Variance, Difference).}
\label{tab:merged_color_jepa}

\begin{tabular}{l c cc cc}
\toprule
\multirow{2}{*}{\textbf{Method}} &
\multirow{2}{*}{\makecell[c]{\textbf{Training}\\\textbf{Free}}} &
\multicolumn{2}{c}{\textbf{Color-shift}} &
\multicolumn{2}{c}{\textbf{JEPA Consistency}} \\
\cmidrule(lr){3-4} \cmidrule(lr){5-6}
& &
\makecell[c]{L1 $\downarrow$} & \makecell[c]{Correlation $\uparrow$} &
\makecell[c]{Variance $\downarrow$} & \makecell[c]{Difference $\downarrow$} \\
\midrule
Rolling Forcing  & \xmark & 0.436 & 0.858 & 0.0162 & 0.201 \\
LongLive         & \xmark & 0.701 & 0.724 & 0.0151 & 0.101 \\
CausVid          & \xmark & 1.047 & 0.451 & 0.0199 & 0.313 \\
\rowcolor{gray!10} \textbf{CV + Ours} & \cmark & 0.607 & 0.778 & 0.0157 & 0.164 \\
Self-Forcing     & \xmark & 1.028 & 0.479 & 0.0145 & 0.191 \\
\rowcolor{gray!10} \textbf{SF + Ours} & \cmark & 0.644 & 0.710 & 0.0108 & 0.170 \\
\bottomrule
\end{tabular}
\end{table}

\begin{figure*}[t]
  \hfill
  \centering
  \includegraphics[width=\linewidth]{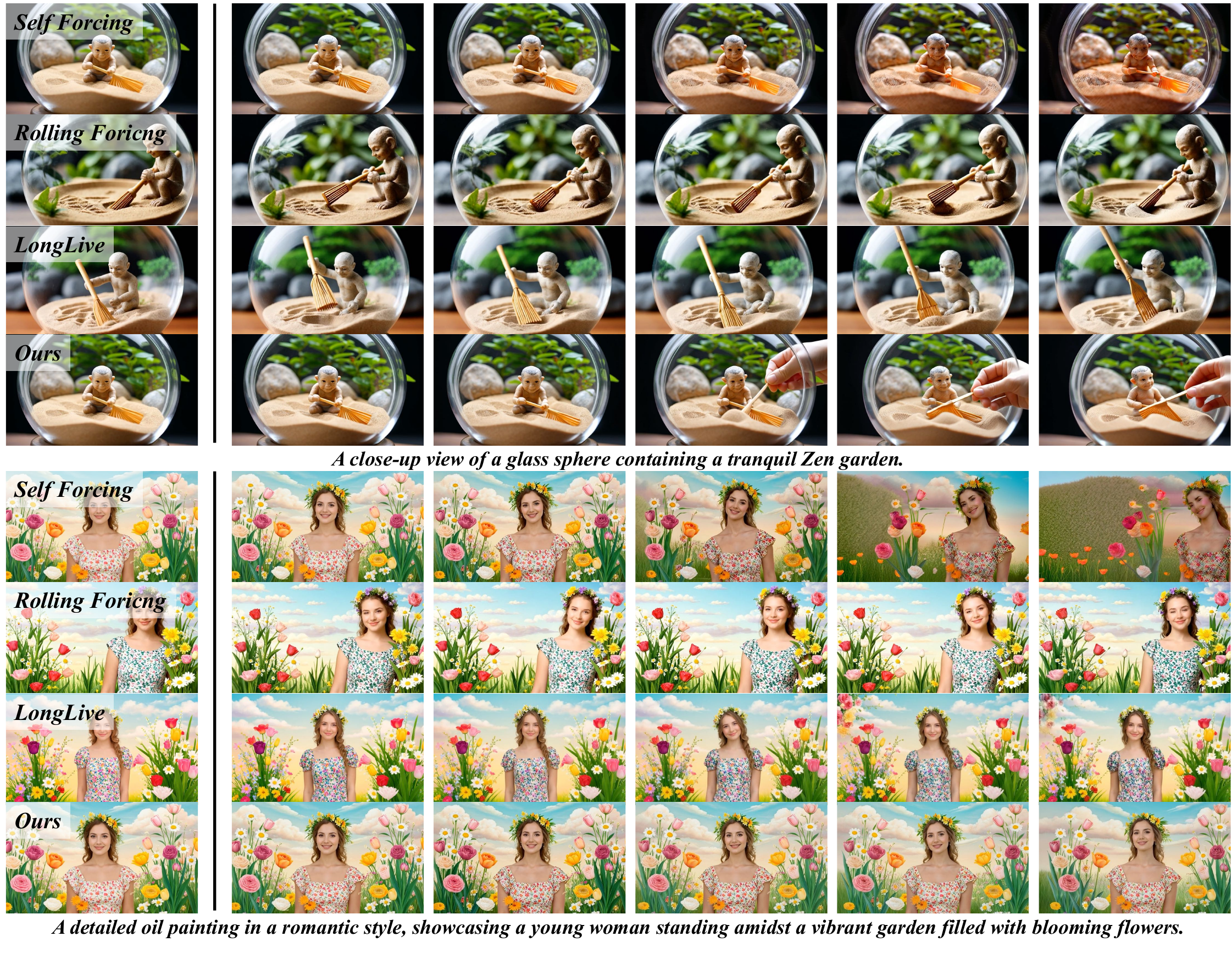}
  \caption{Qualitative comparison of 30-second long-horizon video generation with Self-Forcing, Rolling Forcing, and LongLive. Our method significantly outperforms Self-Forcing and achieves temporal coherence and visual quality comparable to training-based methods.}
  \label{fig:Qualitative}
\end{figure*}

\begin{table}[htbp!]
\scriptsize 
\centering
\setlength{\tabcolsep}{1.5pt} 
\renewcommand{\arraystretch}{1.2}
\caption{\textbf{Comprehensive comparison with test-time scaling methods, including Best-of-N (BoN) and Search-over-Path (SoP), on prompt-conditioned 30-second video generation.}
We report Throughput (fps) and VBench metrics.}
\label{tab:tts_tto}
\begin{tabular}{l c c cccccc}
\toprule
\multirow{2}{*}{\textbf{Method}} &
\multirow{2}{*}{\makecell[c]{\textbf{Train.}\\\textbf{Free}}} & 
\multicolumn{1}{c}{\textbf{Speed}} &
\multicolumn{6}{c}{\textbf{VBench Metrics $\uparrow$}} \\
\cmidrule(lr){3-3} \cmidrule(lr){4-9}
& & \makecell[c]{Total\\fps} &
\makecell[c]{Subject\\Consistency} &
\makecell[c]{Background\\Consistency} &
\makecell[c]{Dynamic\\Degree} &
\makecell[c]{Motion\\Smoothness} &
\makecell[c]{Imaging\\Quality} &
\makecell[c]{Aesthetic\\Quality} \\
\midrule
Self-Forcing      & \xmark & 15.79 & 92.5 & 93.2 & 62.5 & 98.0 & 72.5 & 63.4 \\
SF + BoN          & \cmark &  3.16 & 92.4 & 93.2 & 62.5 & 98.4 & 72.7 & 63.3 \\
SF + SoP          & \cmark &  3.16 &  92.7   &  93.4   &  60.2   &  98.6   &  72.7   &  63.1   \\
\rowcolor{gray!10} \textbf{SF + Ours} & \cmark & 10.53 & 94.0 & 94.2 & 60.2 & 98.3 & 72.7 & 63.8 \\
\bottomrule
\end{tabular}
\end{table}

\section{Experiments}
\subsection{Settings}
\noindent \textbf{Baseline.}
We evaluate our test-time correction method on two baseline models, CausVid~\cite{yin2025slow} and Self-Forcing~\cite{huang2025self}.
Both baselines are built upon the Wan2.1-T2V-1.3B model~\cite{wan2025} and generate 5-second video clips at 16 FPS with a resolution of $832 \times 480$.  More implementation details for fair comparison are included in the supplementary material.

\noindent \textbf{Standard Evaluation.}
We benchmark our method against representative autoregressive video diffusion baselines, including CausVid~\cite{yin2025slow}, Self-Forcing~\cite{huang2025self}, Rolling Forcing~\cite{liu2025rolling}, and LongLive~\cite{yang2025longlive}. Following standard protocols, evaluations are conducted using VBench~\cite{huang2024vbench} on 128 prompts randomly sampled from MovieGen~\cite{polyak2024movie}. Unless otherwise specified, we conduct all experiments in the 30-second video generation setting, using Self-Forcing as the default baseline. More experimental settings and implementation details for fair comparison are included in the supplementary material.

\noindent \textbf{Additional Evaluation.}
To rigorously evaluate temporal coherence, we complement standard VBench quality metrics with temporal color histograms and JEPA scores~\cite{JEPA} to assess long-term temporal drift. Since temporal consistency scores can be artificially improved by suppressing motion, allowing models to effectively cheat the evaluation, we conduct comparisons under matched dynamic degrees to ensure a fair and meaningful assessment. In addition, we use t-LPIPS~\cite{lpips} to explicitly measure visual discontinuities, serving as a direct proxy for flickering artifacts at autoregressive chunk boundaries in our ablation studies. Comprehensive experimental settings and implementation details for fair comparison are included in the supplementary material.

\subsection{Results}
\noindent \textbf{Qualitative Results.}
As shown in Figure~\ref{fig:Qualitative}, integrating our method with Self-Forcing substantially reduces error accumulation in long-horizon video generation.
While the original baselines exhibit temporal drift and visual degradation over time, our integration maintains stable temporal coherence and visual fidelity over 30-second sequences, especially in videos with complex motion and appearance changes.
Under the 30-second setting, our training-free approach achieves visual quality comparable to, and in some cases better than, Rolling Forcing and LongLive, which rely on additional training or specialized mechanisms.
These results demonstrate that our method provides an effective and general test-time solution for improving long-term temporal consistency in autoregressive video generation. More experimental settings and results are included in the supplementary material.

\begin{table}[t]
\centering
\scriptsize
\renewcommand{\arraystretch}{1.1}
\setlength{\tabcolsep}{1.5pt} 
\caption{\textbf{Ablation study on noise-correction steps.} We evaluate quality using VBench metrics alongside the Boundary metric.}
\label{tab:ablation_timesteps}
\begin{tabular}{c ccc cccccc c} 
\toprule
\multirow{3}{*}{\makecell[c]{Total\\NFE}} &
\multicolumn{3}{c}{\textbf{Timesteps}} & 
\multicolumn{6}{c}{\textbf{VBench Metrics $\uparrow$}} &
\multicolumn{1}{c}{\textbf{Boundary $\downarrow$}} \\ 
\cmidrule(lr){2-4} \cmidrule(lr){5-10} \cmidrule(lr){11-11}
 & 750 & 500 & 250 & 
   \makecell[c]{Subject\\Consistency} &
    \makecell[c]{Background\\Consistency} &
    \makecell[c]{Dynamic\\Degree} &
    \makecell[c]{Motion\\Smoothness} &
    \makecell[c]{Imaging\\Quality} &
    \makecell[c]{Aesthetic\\Quality} &
   \makecell[c]{t-LPIPS} \\ 
\midrule
4 & \xmark & \xmark & \xmark & 92.5 & 93.2 &62.5  &98.0  &72.5  &63.4 & 0.178 \\
\midrule
5 & \cmark & \xmark & \xmark & 93.6 & 94.3 & 60.2 & 98.6 & 72.6 & 63.2 & 0.161 \\
5 & \xmark & \cmark & \xmark & 93.2 & 93.9 & 60.9 & 98.5 & 72.8 & 63.1 & 0.182 \\
5 & \xmark & \xmark & \cmark & 93.6 & 94.1 & 57.0 & 98.5 & 72.9 & 63.4 & 0.183 \\
\midrule
\rowcolor{gray!10} 6 & \xmark & \cmark & \cmark & 94.0 & 94.2 & 60.2 & 98.3 & 72.7 & 63.8 & 0.176 \\
6 & \cmark & \cmark & \xmark & 93.1 & 93.9 & 61.7 & 98.4 & 73.0 & 63.1 &  0.170\\
\midrule
7 & \cmark & \cmark & \cmark & 93.4& 94.2 & 62.5 & 98.5 & 72.4 & 63.8 & 0.169 \\
\bottomrule
\end{tabular}
\end{table}

\noindent \textbf{Quantitative Results.}
All quantitative results are summarized in Table~\ref{tab:merged_speed_vbench} and Table~\ref{tab:merged_color_jepa}. Under the 30-second generation setting on VBench, integrating our method into standard autoregressive baselines, Self-Forcing and CausVid, consistently improves long-horizon video generation quality across diverse prompts and scenes. In particular, our method substantially reduces error accumulation and temporal drift, leading to improved subject and background consistency while notably enhancing dynamic degree without sacrificing motion smoothness or imaging quality. Moreover, the proposed path-wise correction effectively stabilizes appearance evolution over time, as evidenced by lower color-shift L1 distances and higher histogram correlations between the first and last frames. At the semantic level, our method also improves JEPA consistency by reducing both the standard deviation and first–last score difference across the entire sequence, indicating more coherent long-term representations. Compared with training-based methods such as Rolling Forcing and LongLive, our approach achieves comparable long-horizon consistency and visual quality while preserving stronger motion dynamics and requiring no additional training or parameter updates at test time.

\noindent \textbf{Comparison with Test-Time Scaling.}
We benchmark our approach against test-time scaling strategies, including Best-of-N (BoN) and Search-over-Path (SoP), as shown in Table~\ref{tab:tts_tto}. While these methods attempt to mitigate errors through redundant candidate generation or iterative search, they incur prohibitive computational overhead and inference latency. In contrast, our method embeds correction directly into a single stochastic sampling trajectory. This design drastically reduces inference costs compared to multi-sample scaling, and by \textit{actively} rectifying structural deviations rather than \textit{passively} selecting from drifting candidates, it achieves superior suppression of long-term error accumulation with minimal overhead. More experimental settings and implementation details for are included in the supplementary material.

\begin{figure}[t]
\centering

\begin{minipage}{0.48\columnwidth}
    \centering
    \includegraphics[width=\linewidth]{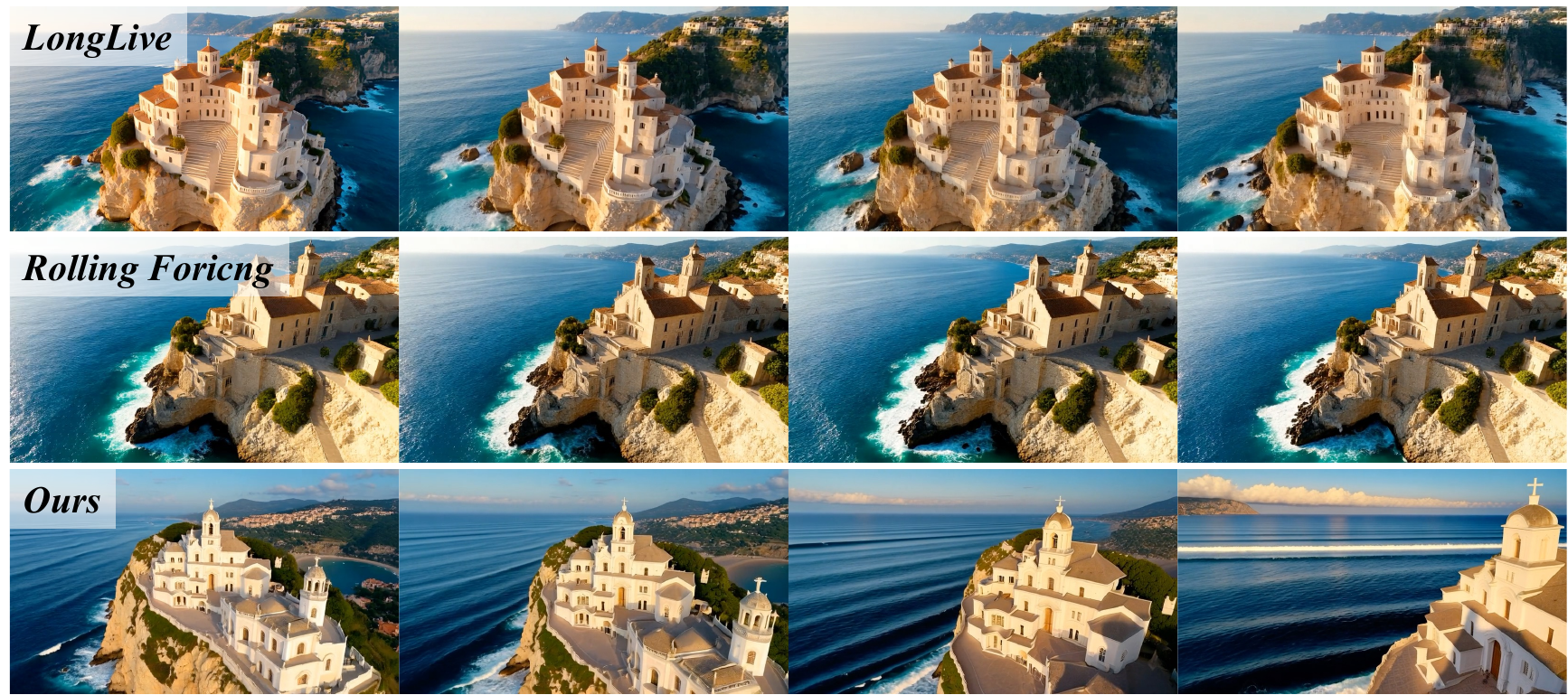}
    \captionof{figure}{\textbf{Comparison between the sink-based method and path-wise correction.} 
    The sink-based method overly constrains intermediate states, leading to degraded motion dynamics and reduced temporal variation.}
    \label{fig:sink}
\end{minipage}
\hfill
\begin{minipage}{0.48\columnwidth}
    \centering
    \includegraphics[width=\linewidth]{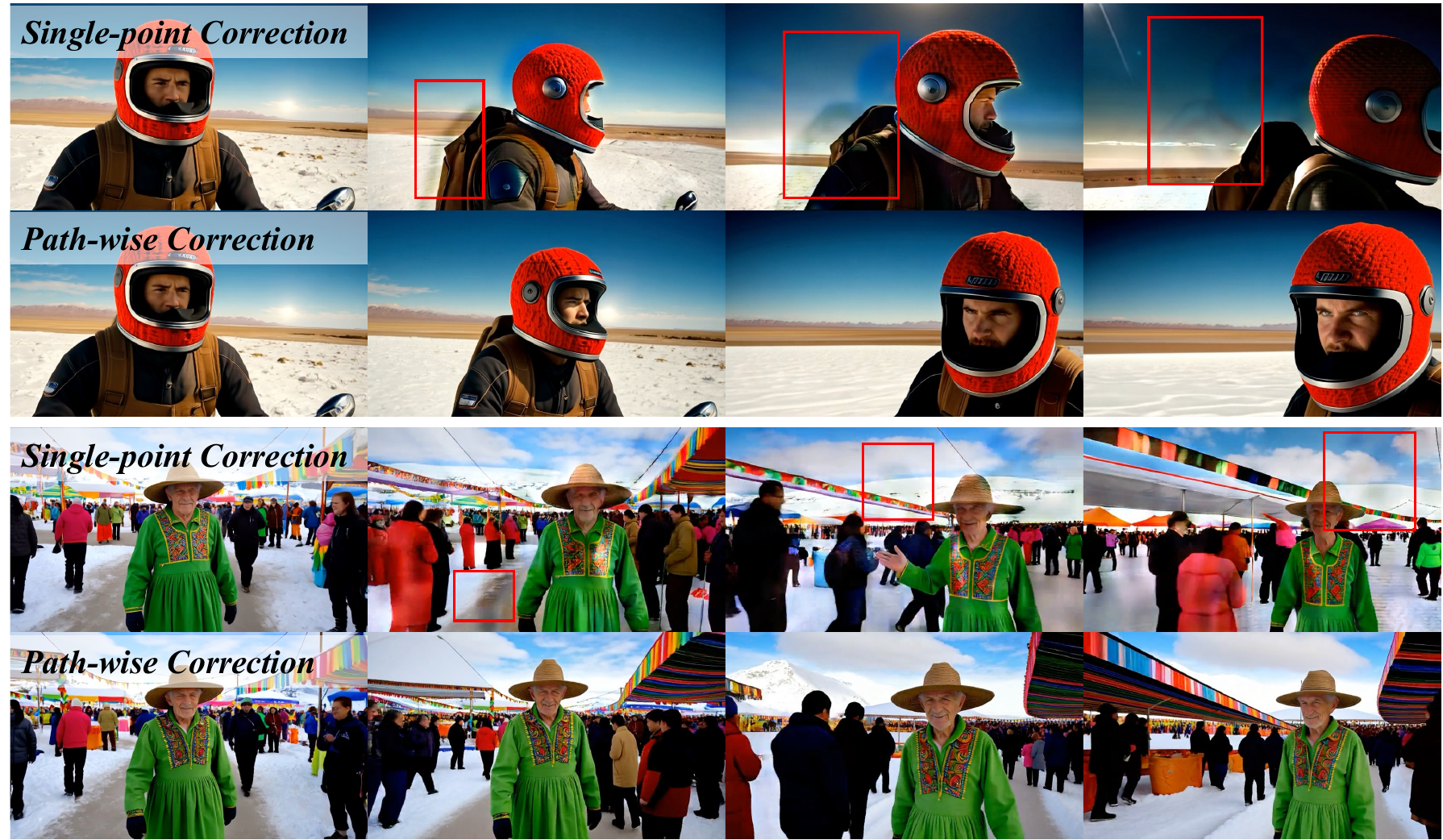}
    \captionof{figure}{\textbf{Comparison of single-point and path-wise correction.} 
    Single-point correction causes temporal discontinuities, while on-path re-noising improves temporal stability and reduces flickering.}
    \label{fig:singlepoint}
\end{minipage}

\end{figure}

\noindent \textbf{Ablation Study on Pathwise Correction.}
We compare \emph{single-point} and \emph{path-wise} correction to evaluate the role of the stochastic sampling trajectory in practice. Single-point correction directly replaces the latent at a fixed denoising step, whereas path-wise correction re-noises the corrected prediction to the same noise level and resumes denoising along the original trajectory. As shown in Figure~\ref{fig:singlepoint} and Table~\ref{tab:correction_strategy}, single-point correction frequently introduces flickering and temporal instability, leading to degraded consistency metrics and higher t-LPIPS scores. In contrast, path-wise correction achieves consistently higher temporal consistency and substantially lower t-LPIPS, resulting in more stable videos with improved temporal coherence. These results demonstrate that effective test-time intervention requires integrating corrections along the sampling path rather than directly replacing latents.

\noindent \textbf{Ablation Study on Noise-correction Steps.}
After establishing the necessity of path-wise correction, we evaluate how different numbers and placements of correction steps along the sampling path affect long-horizon generation quality as shown in Table~\ref{tab:ablation_timesteps}. Enabling correction at noise levels 750, 500, and 250, either individually or in combination, consistently outperforms the baseline without correction, demonstrating the robustness and effectiveness of our method across different configurations. 
Balancing generation quality and inference overhead, we choose correction at noise levels 500 and 250 as our default, since this configuration yields consistent improvements in both quantitative metrics and qualitative visual fidelity.



\noindent \textbf{Comparison with the Sink-based Method.}
We compare our proposed path-wise correction with the Sink-based method. The Sink-based approach keeps the Sink frame as visible context throughout the entire denoising process, effectively imposing persistent conditioning. In contrast, path-wise correction is applied only at later stages after structural stabilization, where corrected predictions are re-noised and integrated along the stochastic sampling trajectory.
Because the Sink frame continuously participates in all denoising steps, the model becomes overly conditioned on it, causing generated content to remain visually and structurally close to the Sink frame. This static conditioning restricts motion and scene variation, suppressing temporal dynamics, as shown in Table~\ref{tab:merged_speed_vbench} and Figure~\ref{fig:sink}. By contrast, path-wise correction preserves structural flexibility in early stages and introduces correction only during appearance refinement, maintaining temporal coherence while retaining meaningful video dynamics.

\begin{table}[t]
\centering
\scriptsize
\setlength{\tabcolsep}{1.5pt} 
\renewcommand{\arraystretch}{1.2}
\caption{\textbf{Comparison of correction strategies, including single-point correction and pathwise correction.} We evaluate quality using VBench metrics alongside the Boundary metric.}
\label{tab:correction_strategy}
\begin{tabular}{l cccccc c}
\toprule
\multirow{2}{*}{\textbf{Method}} &
\multicolumn{6}{c}{\textbf{VBench Metrics $\uparrow$}} &
\multicolumn{1}{c}{\textbf{Boundary $\downarrow$}} \\
\cmidrule(lr){2-7} \cmidrule(lr){8-8}
   & \makecell[c]{Subject\\Consistency} &
    \makecell[c]{Background\\Consistency} &
    \makecell[c]{Dynamic\\Degree} &
    \makecell[c]{Motion\\Smoothness} &
    \makecell[c]{Imaging\\Quality} &
    \makecell[c]{Aesthetic\\Quality} &
   \makecell[c]{t-LPIPS} \\ 
\midrule
Single-point & 93.4 & 94.0 & 57.0 & 98.3 & 71.6 & 62.8 & 0.205 \\ 

\rowcolor{gray!10} \textbf{Pathwise} & 94.0 & 94.2 & 60.2 & 98.3 & 72.7 & 63.8 & 0.176 \\
\bottomrule
\end{tabular}
\end{table}

\begin{table}[t!]
\centering
\scriptsize
\setlength{\tabcolsep}{1.5pt} 
\renewcommand{\arraystretch}{1.2}
\caption{\textbf{Comparison on prompt-conditioned 5-second video generation.} We evaluate quality using standard VBench metrics.}
\label{tab:short_results}

\begin{tabular}{l cccccc}
\toprule
\multirow{2}{*}{\textbf{Method}} &
\multicolumn{6}{c}{\textbf{VBench Metrics $\uparrow$}} \\
\cmidrule(lr){2-7}
 & \makecell[c]{Subject\\Consistency} &
    \makecell[c]{Background\\Consistency} &
    \makecell[c]{Dynamic\\Degree} &
    \makecell[c]{Motion\\Smoothness} &
    \makecell[c]{Imaging\\Quality} &
    \makecell[c]{Aesthetic\\Quality} \\
\midrule

CausVid & 96.2 & 94.9 & 54.7 & 98.2 & 70.5 & 63.8 \\
\rowcolor{gray!10} \textbf{CausVid + Ours} & \textbf{96.6} & \textbf{95.2} & \textbf{68.0} & {97.8} & \textbf{70.5} & \textbf{64.2} \\
\midrule
Self-Forcing & 97.0 & 96.2 & 62.5 & 98.7 & 72.9 & 64.5 \\
\rowcolor{gray!10} \textbf{Self-Forcing + Ours} & \textbf{97.0} & \textbf{96.3} & \textbf{62.5} & \textbf{98.7} & \textbf{73.0} & \textbf{64.6} \\

\bottomrule
\end{tabular}
\end{table}

\noindent \textbf{Comparison on Short Video Generation.}
As shown in Table~\ref{tab:short_results}, we evaluate our method on short video generation. Although error accumulation is less pronounced under short temporal horizons, our method still consistently outperforms the baseline across most metrics. This indicates that the proposed correction strategy is not specialized to long-horizon generation, but also remains effective in short video settings. Together with the significant improvements observed for long video generation, these results demonstrate the robustness and general applicability of our approach.

\section{Conclusion}
In this paper, we propose Test-Time Correction, a training-free test-time method for stabilizing distilled autoregressive diffusion models in long-horizon video generation. The proposed approach addresses error accumulation by introducing training-free, reference-based correction along the stochastic sampling process, allowing corrected predictions to be smoothly inherited by subsequent denoising steps. Without modifying model parameters or requiring additional training, our method effectively suppresses temporal drift while preserving the original generation behavior. Extensive experiments demonstrate that Test-Time Correction consistently improves long-horizon stability across multiple distilled video generation models, extending the achievable generation length to 30 seconds with a slight computational overhead and competitive visual quality.

%
%
\bibliographystyle{splncs04}
\bibliography{main}

\newpage
\appendix

\section{Details on Samplers.}
\noindent \textbf{Few-step stochastic sampling.}
The autoregressive video diffusion backbones used in our framework—\emph{Self-Forcing} and \emph{CausVid}—are obtained by step distillation from a multi-step \emph{bidirectional} video diffusion model trained with the \emph{Rectified Flow} (RF) objective.
In RF, the forward noising process is defined by a linear interpolation between a clean latent video $x_0$ and an isotropic Gaussian terminal state $x_{T_{\max}} \sim \mathcal{N}(0,I)$:
\begin{equation}
x_t = t\,x_0 + (1-t)\,x_{T_{\max}}, \qquad t \in [0,1].
\label{eq:rf_forward_interp}
\end{equation}
Differentiating \eqref{eq:rf_forward_interp} with respect to $t$ gives the corresponding velocity along the path,
\begin{equation}
\mathbf{v}_t \triangleq \frac{\mathrm{d}x_t}{\mathrm{d}t} = x_0 - x_{T_{\max}},
\label{eq:rf_gt_velocity}
\end{equation}
which is constant in $t$ under this parameterization.
A time-conditioned flow network $v_{\theta_0}(x_t,t)$ is trained to regress this velocity via mean squared error,
\begin{equation}
\mathcal{L}_{\text{flow}}
=
\mathbb{E}_{x_0,\,x_{T_{\max}},\,t}
\Big[
\big\|
v_{\theta_0}(x_t,t) - \mathbf{v}_t
\big\|_2^2
\Big].
\label{eq:rf_loss}
\end{equation}

For long-horizon video generation, the bidirectional RF predictor $v_{\theta_0}$ is distilled into a \emph{causal} autoregressive model $v_\theta$ by replacing bidirectional attention with causal attention, so that the prediction for the $i$-th frame conditions only on previously generated frames $x^{<i}$ (with KV caching used to reuse past attention states during sequential generation).
At inference, sampling is carried out on a small set of discrete timesteps $\{T_J > T_{J-1} > \cdots > T_0\}$ with $J$ much smaller than standard multi-step samplers.
Given a noisy latent at step $T_j$, the model outputs a denoised estimate using the RF update form
\begin{equation}
\hat{x}^{\,i}_{0 \mid t_j}
=
G_\theta\!\left(x^{\,i}_{t_j};\,x^{<i},t_j\right)
=
x^{\,i}_{t_j} + (1-t_j)\,v_\theta\!\left(x^{\,i}_{t_j};\,x^{<i},t_j\right),
\label{eq:tweedie_form}
\end{equation}
and then constructs the next noisy state $x^{\,i}_{t_{j-1}}$ by re-applying the RF forward interpolation with newly sampled Gaussian noise, i.e., using \eqref{eq:rf_forward_interp} with $t=t_{j-1}$.
This yields a stochastic few-step sampler in which independent Gaussian noise is injected at each transition between adjacent timesteps.

\noindent \textbf{ODE-based sampling.}
Rectified Flow also supports a deterministic sampler by treating the learned velocity predictor as an ordinary differential equation (ODE).
Given the causal autoregressive velocity field $v_\theta(\cdot;\,x^{<i},t)$, one can define the sampling dynamics as
\begin{equation}
\frac{\mathrm{d} x_t}{\mathrm{d} t}
=
v_\theta(x_t;\, x^{<i}, t),
\qquad
x_{t=1} \sim \mathcal{N}(0, I),
\label{eq:rf_ode}
\end{equation}
which maps an initial Gaussian state at $t=1$ to a terminal sample at $t=0$ through deterministic integration.

In practice, \eqref{eq:rf_ode} is approximated on a discrete time grid $\{T_J > T_{J-1} > \cdots > T_0\}$.
Compared with the stochastic transition that re-samples Gaussian noise at each step,
\begin{equation}
f_{\theta,t_j}\!\left(x^{\,i}_{t_j}\right)
=
\Psi\!\left(
\hat{x}^{\,i}_{0 \mid t_j},
\epsilon^{\,i}_{j-1},
t_{j-1}
\right),
\qquad
\epsilon^{\,i}_{j-1} \sim \mathcal{N}(0, I),
\label{eq:stochastic_step}
\end{equation}
an ODE sampler removes the noise variable $\epsilon^{\,i}_{j-1}$ and replaces the transition with a deterministic one-step integrator.
Using the explicit Euler method, the update from $t_j$ to $t_{j-1}$ is
\begin{equation}
\tilde{f}_{\theta,t_j}\!\left(x^{\,i}_{t_j}\right)
=
x^{\,i}_{t_j}
+
(t_{j-1} - t_j)\,
v_\theta\!\left(
x^{\,i}_{t_j};\, x^{<i}, t_j
\right).
\label{eq:ode_step}
\end{equation}
Equation \eqref{eq:ode_step} is the standard explicit Euler discretization of the ODE \eqref{eq:rf_ode} on the chosen timestep schedule.

Finally, note that the two samplers differ only in whether the transition between adjacent timesteps introduces an additional Gaussian perturbation (stochastic) or performs a purely deterministic numerical integration step (ODE).
Under the ODE formulation, once the initial state $x_{t=1}$ and the discretization scheme are fixed, the generated trajectory is fully determined by repeated application of \eqref{eq:ode_step}.

\section{Details on Evaluations}
\label{appendix:evaluation_details}

\noindent \textbf{Boundary Continuity (t-LPIPS).}
To measure perceptual discontinuities at segment junctions in autoregressive generation, we compute LPIPS only on \emph{boundary-adjacent} frame pairs.
Our autoregressive model generates a video as $K$ consecutive chunks.
Let $t_k$ denote the last frame index of the $k$-th chunk; then the $k$-th boundary is the adjacent pair $(f_{t_k}, f_{t_k+1})$ for $k\in\{1,\ldots,K-1\}$.
We define the boundary score as the mean LPIPS over these $K-1$ pairs:
\begin{equation}
\text{LPIPS}_{\text{boundary}}
=
\frac{1}{K-1}\sum_{k=1}^{K-1}\text{LPIPS}\!\left(f_{t_k},\,f_{t_k+1}\right).
\end{equation}
This metric isolates changes that occur specifically when switching from one generated chunk to the next, rather than averaging over all within-chunk frame pairs.

\noindent \textbf{Color Shift (HSV Histogram).}
To quantify color distribution changes across the generated sequence, we compare the color histograms of the first and last frames.
Let $f_{\text{start}}$ and $f_{\text{end}}$ be the initial and final frames of a generated video.
We convert both frames to HSV space and compute an $L_1$-normalized histogram of the Hue channel using 180 bins, denoted by $h_{\text{start}}, h_{\text{end}} \in \mathbb{R}^{180}$ with $\|h_{\text{start}}\|_1=\|h_{\text{end}}\|_1=1$.
We report two statistics between $h_{\text{start}}$ and $h_{\text{end}}$:
(i) the $L_1$ distance, $\|h_{\text{start}}-h_{\text{end}}\|_1$, and
(ii) the Pearson correlation coefficient, $\rho(h_{\text{start}}, h_{\text{end}})$.

\begin{figure}[h]
\begin{wrapcode}

import torch
import torch.nn.functional as F
from torch.autograd.functional import jacobian

# phi(.) : frozen JEPA-style encoder (e.g., V-JEPA / DINOv2)

# (1) JEPA-score: Jacobian-based local complexity / density proxy
J = jacobian(lambda x: phi(x).sum(0), X)          # X: (B,C,H,W)

with torch.inference_mode():
    J = J.flatten(2).permute(1, 0, 2)             # (B, d, HW)
    sv = torch.linalg.svdvals(J)                  # singular values
    JEPA_score = sv.clamp(min=eps).log().sum(1)   # (B,)

# (2) JEPA consistency: first-frame anchored temporal drift
Z = phi(I_1_T)                                    # (T, d)
Z = F.normalize(Z, dim=-1)
z_ref = Z[0]
d_t = 1.0 - (Z @ z_ref)                           # (T,)

JEPA_Std  = d_t.std()
JEPA_Diff = (d_t[-1] - d_t[0]).abs()
\end{wrapcode}
\caption{\textbf{JEPA-based long-video evaluation pseudocode.} Given frames $I_{1:T}$ and a frozen encoder $\phi(\cdot)$, we compute JEPA-score from $\mathrm{svd}(\nabla_X \phi(X))$ and temporal drift statistics from first-frame-anchored feature cosine distances.}
\label{lst:jepa_full}
\end{figure}








\noindent \textbf{JEPA Consistency} To rigorously quantify both the intrinsic distribution fidelity and the long-horizon semantic stability of autoregressive video generation, we adopt a dual-metric evaluation framework based on a frozen V-JEPA encoder $\phi(\cdot)$, grounded in recent theoretical findings that JEPA representations implicitly encode data density through Gaussian embeddings and local volume changes of the encoder mapping. Specifically, for each generated frame (or short temporal clip) $x_t$, we compute the encoder Jacobian $J_\phi(x_t)=\partial \phi(x_t)/\partial x_t$ and define an \emph{Intrinsic Density Score} as $S_t^{\mathrm{dens}} \propto \frac{1}{2}\log\det\!\left(J_\phi(x_t)^\top J_\phi(x_t)\right)$, which estimates the local log-volume expansion induced by $\phi$ and thus serves as a proxy for the sample's likelihood under the learned data manifold; a monotonic decay of $S_t^{\mathrm{dens}}$ along time indicates progressive manifold departure and hallucination as the generation drifts into low-density regions of the data distribution. In parallel, to measure global semantic consistency, we compute the normalized embedding trajectory $\mathbf{z}_t=\mathrm{norm}(\phi(x_t))$ and define the \emph{Temporal Drift Distance} relative to the initial semantic anchor as $d_t = 1 - \mathbf{z}_t^\top \mathbf{z}_1$, which captures distributional deviation in the JEPA-induced representation space. Aggregating these frame-wise measurements at a fixed temporal granularity (e.g., per second), we report two summary statistics: $\mathrm{JEPA\text{-}Std}=\mathrm{Std}(\{d_t\}_{t=1}^{T})$ to characterize the volatility of representation drift, and $\mathrm{JEPA\text{-}Diff}=|d_T-d_1|$ to quantify the accumulated long-range semantic deviation, thereby providing a holistic assessment of a model’s robustness to both distributional collapse and semantic drift in long-horizon video generation.

\noindent \textbf{Test-time Scaling Configuration.}
We compare against two inference-time scaling protocols under a fixed sampling budget of $N=5$.
\textit{Best-of-$N$ (BoN)} performs selection at the \emph{trajectory} level.
For each video segment, we run $N$ independent sampling trajectories by drawing $N$ independent initial noise latents.
Each trajectory is rolled out to a complete segment, and we compute a scalar reward for the resulting segment.
Among the $N$ completed candidates, we keep the one with the highest reward score as the output of that segment.
\textit{Search-over-Path (SoP)} performs selection at the \emph{step} level on the same timestep schedule.
At each denoising timestep, we generate $N$ candidate next-step latents by injecting $N$ independent Gaussian noise realizations for that transition (equivalently, $N$ candidate stochastic updates from the current latent).
We then evaluate the reward for each candidate at that timestep and select the candidate with the highest reward as the current latent for the next timestep.
This greedy selection is repeated until the segment is completed.

\section{Details on Methods.}
\noindent \textbf{Details on Test-time Optimization (TTO).} Following the test-time adaptation protocols established in HyperNoise and AutoRefiner, we perform gradient-based optimization at each sampling step. We employ an AdamW optimizer with a learning rate of $1 \times 10^{-4}$. Specifically, at each denoising step for each latent chunk, the latent prediction is first decoded into pixel space via a pre-trained VAE decoder. We then compute the Mean Squared Error (MSE) loss and the CLIP score on the decoded image with the initial image, which serve as proxies for pixel-level and semantic-level rewards, respectively, to guide the optimization process.

\section{Further Quantitative Results.}
\noindent \textbf{Full VBench Scores.} We conduct a comprehensive evaluation on the full VBench benchmark, using
all 946 prompts and covering all 16 metrics reported in Table~\ref{tab:vbench_full}. For detailed metric definitions, we refer readers to the VBench paper. All values are computed with the official standardized
evaluation scripts. Our method achieves substantial improvements in overall quality, particularly in
frame-wise fidelity, and also outperforms distilled baselines on semantic scores.

\begin{table}[h!]
\centering
\scriptsize
\renewcommand{\arraystretch}{1.3}
\caption{\textbf{Full evaluation on VBench metrics.} We evaluate the performance across all quality and semantic dimensions.}
\label{tab:vbench_full}

\resizebox{\linewidth}{!}{
    \setlength{\tabcolsep}{4pt} 
    \begin{tabular}{l | ccccccc | c}
    \toprule
    \multirow{3}{*}{\textbf{Method}} & \multicolumn{7}{c|}{\textbf{Quality Metrics}} & \multirow{3}{*}{\makecell[c]{\textbf{Quality}\\\textbf{Score}}} \\
    \cmidrule(lr){2-8}
    & \makecell[c]{Subject\\Consistency} & \makecell[c]{Background\\Consistency} & \makecell[c]{Temporal\\Flickering} & \makecell[c]{Motion\\Smoothness} & \makecell[c]{Dynamic\\Degree} & \makecell[c]{Aesthetic\\Quality} & \makecell[c]{Imaging\\Quality} & \\
    \midrule
    CausVid &89.1&90.7&99.3&98.0&62.5&61.7&65.3&80.8 \\
    \rowcolor{gray!10} \textbf{CausVid + Ours} &91.4&92.2&99.2&97.4&71.9&61.1&66.4&81.9 \\
    \midrule
    Self-Forcing &89.8&90.7&98.1&98.5&69.4&60.0&68.7&81.4 \\
    \rowcolor{gray!10} \textbf{Self-Forcing + Ours} &91.1&91.7&98.2&98.8&68.1&60.7&68.6&82.1 \\
    \bottomrule
    \end{tabular}
}


\resizebox{\linewidth}{!}{
    \setlength{\tabcolsep}{3pt} 
    \begin{tabular}{l | ccccccccc | c}
    \toprule
    \multirow{3}{*}{\textbf{Method}} & \multicolumn{9}{c|}{\textbf{Semantic Metrics}} & \multirow{3}{*}{\makecell[c]{\textbf{Semantic}\\\textbf{Score}}} \\
    \cmidrule(lr){2-10}
    & \makecell[c]{Object\\Class} & \makecell[c]{Multiple\\Objects} & \makecell[c]{Human\\Action} & Color & \makecell[c]{Spatial\\Relationship} & Scene & \makecell[c]{Temporal\\Style} & \makecell[c]{Appearance\\Style} & \makecell[c]{Overall\\Consistency} & \\
    \midrule
    CausVid &77.4&58.8&77.0&84.2&61.8&32.2&22.4&19.9&23.0&65.9 \\
    \rowcolor{gray!10} \textbf{CausVid + Ours} &76.0&62.0&80.0&80.6&63.9&34.9&22.1&19.7&22.9&66.3 \\
    \midrule
    Self-Forcing &81.9&61.9&81.0&88.0&79.2&32.2&23.6&19.6&23.8&70.0 \\
    \rowcolor{gray!10} \textbf{Self-Forcing + Ours} &81.6&66.5&82.0&92.2&80.2&30.8&23.1&19.4&23.6&70.7 \\
    \bottomrule
    \end{tabular}
}
\end{table}

\begin{wraptable}{r}{0.5\textwidth}
    \centering
    \scriptsize
    \setlength{\tabcolsep}{4pt} 
    \vspace{-20pt} 
    
    \caption{\textbf{Dynamic Degree Analysis.}} 
    \label{tab:dynamic_results} 

    \begin{tabular}{l | ccc}
    \toprule
    \multirow{3}{*}{\textbf{Method}} & \multicolumn{3}{c}{\textbf{Dynamic Degree}} \\
    \cmidrule(lr){2-4}
    & LPIPS$\uparrow$ & SSIM$\downarrow$ & PSNR$\downarrow$ \\
    \midrule
    Rolling Forcing & 0.2956 & 0.5738 & 16.2365 \\
    Longlive      & 0.3056 & 0.5969 & 16.8669 \\
    Self-Forcing      & 0.3548 & 0.5377 & 15.6360 \\
    \rowcolor{gray!10} \textbf{Ours}      & 0.3489 & 0.5440 & 15.5279\\
    \bottomrule
    \end{tabular}
    \vspace{-10pt} 
\end{wraptable}

\noindent \textbf{Dynamic Preservation.} Adhering to the evaluation protocols established in AutoRefiner, we conduct a comparative analysis of the dynamic degree against baseline methods, including Self-Forcing, Rolling Forcing, and Longlive. To rigorously assess the dynamic degree, we quantify the perceptual variation between temporally strided frames utilizing metrics such as LPIPS, SSIM, and PSNR with a fixed sampling interval $k$ (e.g., $k=12$). We model the magnitude of motion and structural evolution over time by computing the average distance $D = \mathbb{E}_t [\mathcal{M}(f_t, f_{t+k})]$, where $\mathcal{M}$ represents the specific metric function (e.g., LPIPS) and $f_t$ denotes the frame at time step $t$. As evidenced in Tab.~\ref{tab:dynamic_results}, unlike baseline approaches that often compromise motion magnitude to ensure stability, our method sustains a superior dynamic degree while preserving temporal coherence, thereby effectively maintaining the vividness of the generated content.

\section{Further Qualitative Results.}
\begin{figure*}[t]
  \hfill
  \centering
  \includegraphics[width=\linewidth]{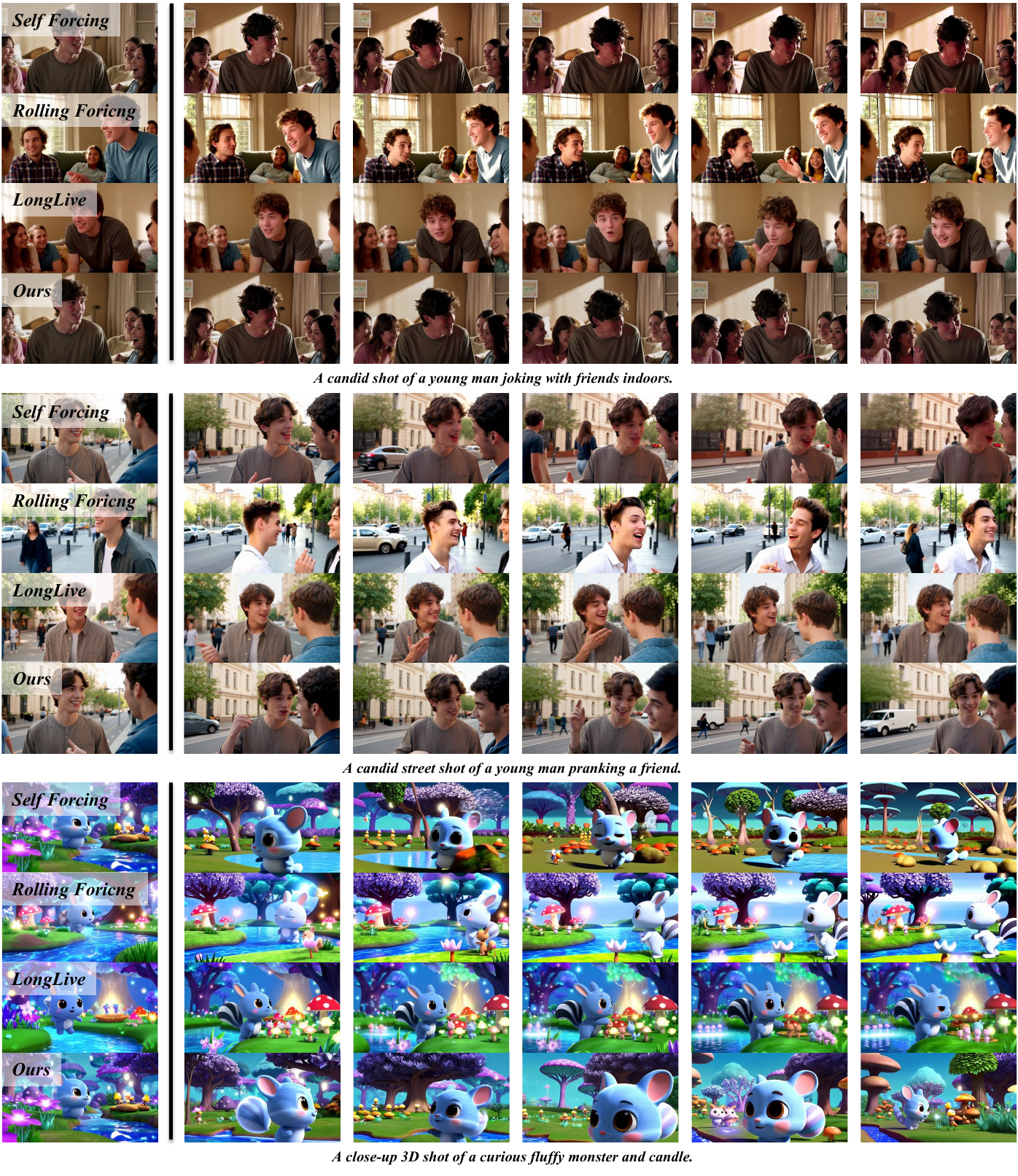}
  \caption{Qualitative comparison of 30-second long-horizon video generation with Self-Forcing, Rolling Forcing, and LongLive. Our method significantly outperforms Self-Forcing and achieves temporal coherence and visual quality comparable to training-based methods.}
  \label{fig:sup_1}
\end{figure*}

\begin{figure*}[t]
  \hfill
  \centering
  \includegraphics[width=\linewidth]{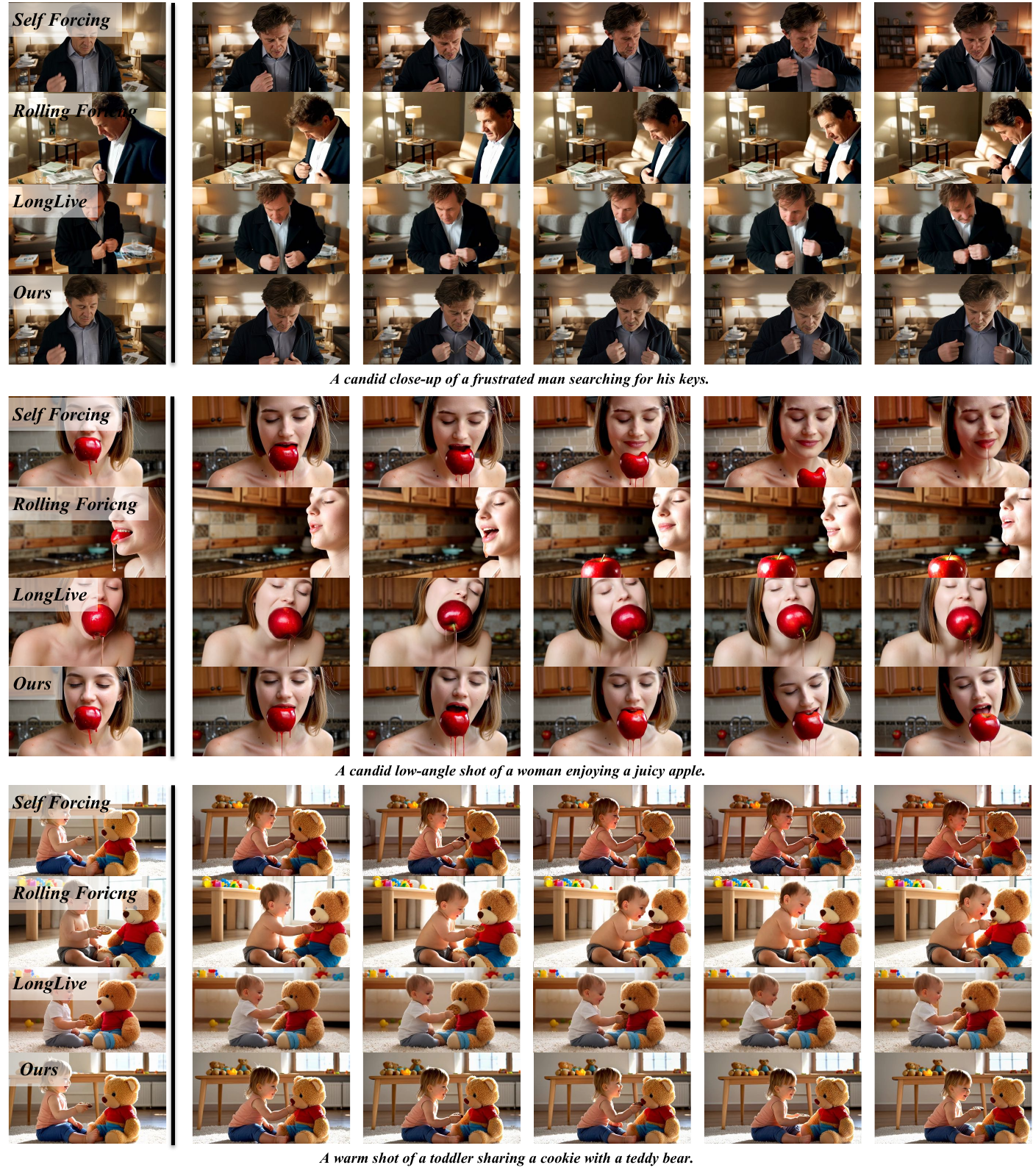}
  \caption{Qualitative comparison of 30-second long-horizon video generation with Self-Forcing, Rolling Forcing, and LongLive. Our method significantly outperforms Self-Forcing and achieves temporal coherence and visual quality comparable to training-based methods.}
  \label{fig:sup_2}
\end{figure*}

\begin{figure*}[t]
  \hfill
  \centering
  \includegraphics[width=\linewidth]{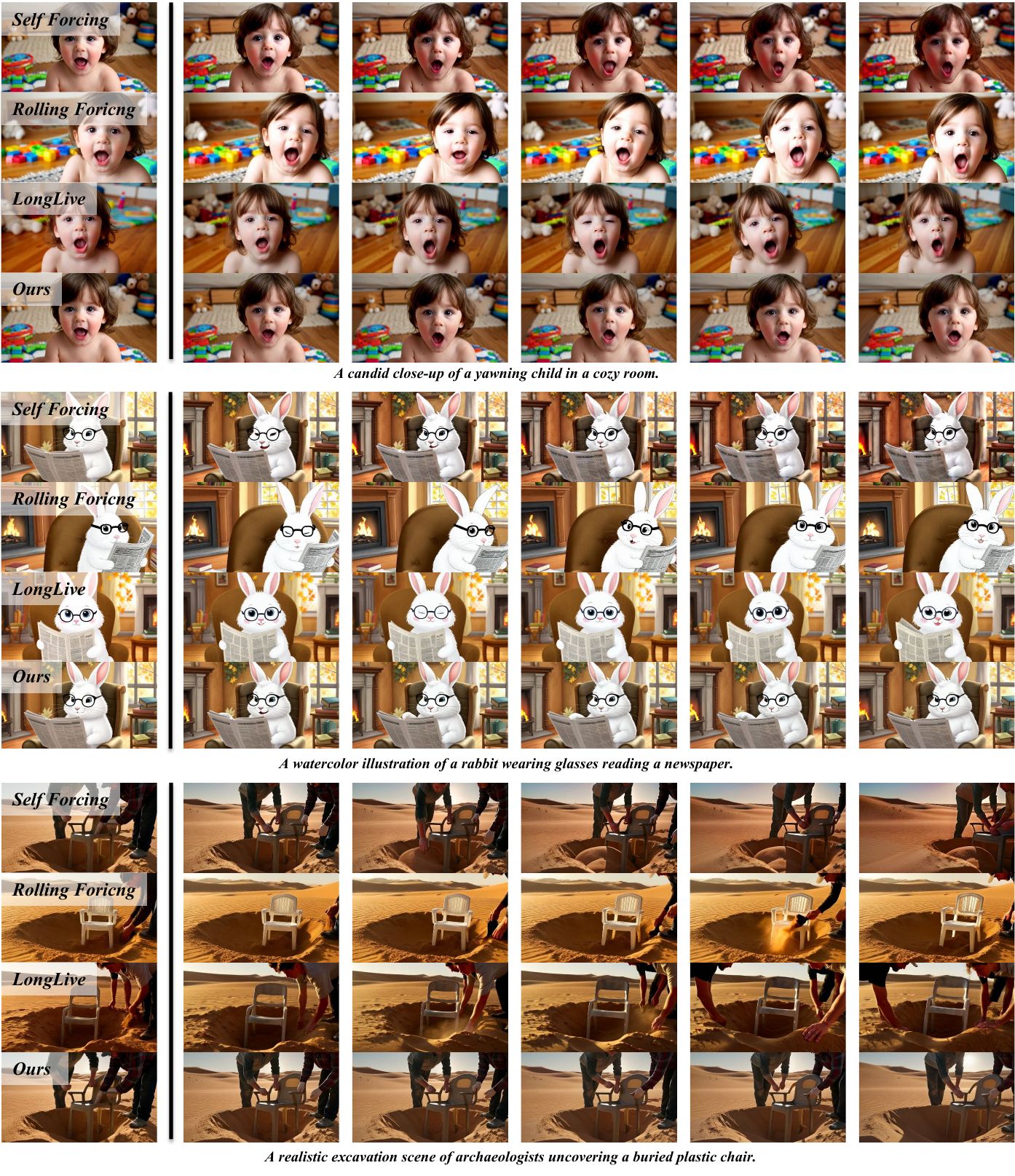}
  \caption{Qualitative comparison of 30-second long-horizon video generation with Self-Forcing, Rolling Forcing, and LongLive. Our method significantly outperforms Self-Forcing and achieves temporal coherence and visual quality comparable to training-based methods.}
  \label{fig:sup_3}
\end{figure*}

We provide additional visual results to further demonstrate the effectiveness of our method. Figure~\ref{fig:sup_1}, Figure~\ref{fig:sup_2}, and Figure~\ref{fig:sup_3} present more generated examples under diverse scenarios. These results consistently exhibit high visual quality and temporal coherence, reinforcing the robustness of our approach across different prompts and settings.

\end{document}